%% file: neurips_2026.tex
\documentclass{article}
\PassOptionsToPackage{numbers}{natbib}

\usepackage[main, final]{neurips_2026}

\usepackage[utf8]{inputenc} 
\usepackage[T1]{fontenc}    
\usepackage{hyperref}       
\usepackage{url}            
\usepackage{booktabs}       
\usepackage{amsfonts}       
\usepackage{nicefrac}       
\usepackage{microtype}      
\usepackage{xcolor}         

\input{math_commands.tex}

\usepackage{float}
\usepackage{hyperref}
\usepackage{url}
\usepackage{algorithm}
\usepackage{graphicx,verbatim}
\usepackage{amssymb}
\usepackage{amsmath}
\usepackage{array}
\usepackage{bm}  
\usepackage{dsfont}
\usepackage{xcolor}
\usepackage{colortbl}
\usepackage{booktabs}
\usepackage{multirow}
\usepackage{makecell}
\usepackage{colortbl}

\def\eg{{\em e.g.}}

\title{Learning Label-Efficient Interpretable Medical Image Diagnosis via Semi-supervised Hypergraph Concept Bottleneck Model}

\author{%
  Yijun Yang\textsuperscript{1,2}\thanks{Equal contribution.}$\quad$
  Ruiqiang Xiao\textsuperscript{1}\footnotemark[1]$\quad$
  Lijie Hu\textsuperscript{3}$\quad$
  Angelica I Aviles-Rivero\textsuperscript{4}$\quad$\\
  \bf
  Yunzhu Wu\textsuperscript{5}$\quad$
  Jing Qin\textsuperscript{6}$\quad$
  Lei Zhu\textsuperscript{1}\thanks{Corresponding author.}\\
  \textsuperscript{1}HKUST(GZ)$\quad$
  \textsuperscript{2}Joy Future Academy$\quad$
  \textsuperscript{3}MBZUAI$\quad$
  \textsuperscript{4}Tsinghua University$\quad$\\
  \textsuperscript{5}Sichuan University$\quad$
  \textsuperscript{6}PolyU$\quad$
}

\begin{document}

\maketitle

\begin{abstract}
Deep learning has revolutionized medical image analysis, delivering exceptional diagnostic accuracy across diverse applications. 
Yet, the lack of interpretability in its decision-making hinders clinical adoption, particularly in high-stakes medical contexts where transparency is paramount for trustworthiness. 
For example, in Placenta Accreta Spectrum (PAS), subtle cues in ultrasound imaging challenge reliable diagnosis, rendering black-box models untrustworthy for accurate scoring.
To address this, Concept Bottleneck Models (CBMs) offer a promising avenue by embedding clinically meaningful intermediate concepts into the diagnosis pipeline, enabling clinicians to scrutinize and refine model outputs. 
However, conventional CBMs falter in capturing complex inter-concept dependencies and demand costly, expert-driven concept annotations, limiting their scalability. 
This study introduces a novel semi-supervised CBM framework designed for medical imaging, which leverages dual-level hypergraph learning to model high-order concept dependencies and generate domain-adaptive pseudo-labels.
Our approach achieves superior interpretability and performance by integrating a concept-level hypergraph for enhanced reasoning and an image-level hypergraph for robust pseudo-label generation. 
Experiments on a newly annotated PAS ultrasound dataset and a breast ultrasound public dataset demonstrate the effectiveness of the proposed concept label-efficient interpretable framework.
Its universality is further validated on the dermoscopic image dataset SkinCon.
The code is available at \url{https://github.com/scott-yjyang/HyperCBM}.
\end{abstract}

\section{Introduction}

Deep learning has driven substantial advancements in medical image analysis, achieving state-of-the-art performance across various diagnostic tasks~\cite{chen2022recent,liu2022deep,zhou2021review,yang2023diffmic,diffmicv2,yang2023mammodg,gong2025cect}.
Yet despite this progress, its clinical adoption remains limited, primarily due to the lack of interpretability. In high-stakes medical decision-making, models must not only be accurate, but also provide transparent reasoning that clinicians can understand, validate, and act upon \cite{tjoa2020survey,reddy2022explainability,reyes2020interpretability,nasarian2024designing,yang2025medical}.
For example, \textit{Placenta Accreta Spectrum (PAS)}, a life-threatening pregnancy complication, demands early and accurate diagnosis due to severe risks like hemorrhage.
This makes interpretability essential for reliable clinical decision-making.
Ultrasound imaging is widely used for PAS considering its non-invasiveness, real-time capability, and cost-effectiveness~\cite{jauniaux2018placenta,cali2019prenatal}.
However, it also presents operator dependency, subtle imaging dynamics, and complex anatomical structures~\cite{sarris2012intra,avola2021ultrasound,yang2025vivim,xu2024lgrnet}, challenging existing deep models.
These facts encourage deep models not only to achieve great results, but also to offer interpretable insights that align with clinical reasoning, enabling trustworthy decisions and intervention in real-world applications.
\begin{figure}[t]
    \centering
    \includegraphics[width=0.9 \textwidth]{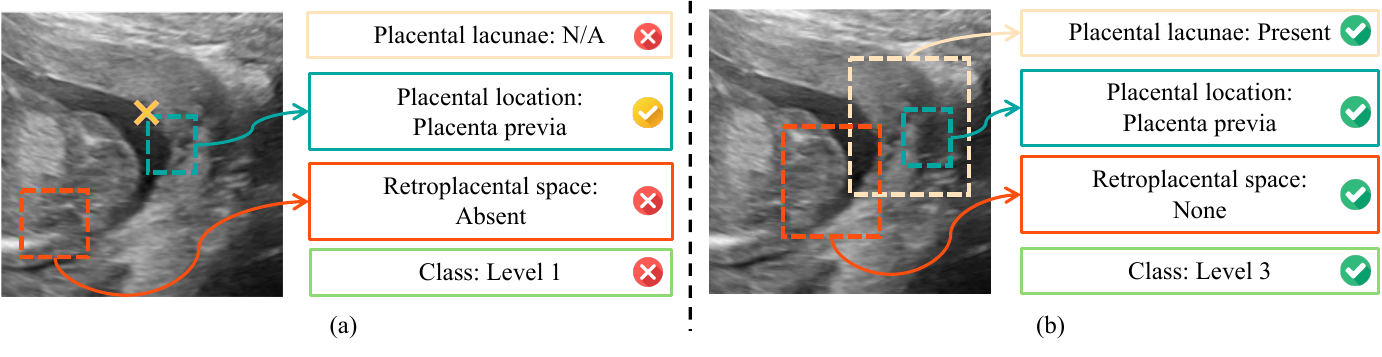}
    \vspace{-2mm}
    \caption{\textbf{Traditional methods degenerate in a semi-supervised spirit.} 
    The conventional CEM (a) and our HyperCBM (b) try to infer the PAS severity level from the predicted concepts. 
    CEM illustrates three error modes: ignoring lacunae, misinterpreting the retroplacental space, and focusing on a biased placental location. These concept errors yield the wrong severity. Instead, HyperCBM successfully predicts severity from the correct concepts and attentions that match expert assessment.}
    \vspace{-4mm}
    \label{teaser}
\end{figure}

Concept Bottleneck Models (CBMs)\cite{cbm,yuksekgonulpost} provide a promising solution by introducing human-understandable intermediate concepts, allowing clinicians to trace decision pathways \ 
and correct mispredictions~\cite{kim2023concept,pang2024integrating,chowdhury2024adacbm}. \ 
However, existing CBMs face two bottlenecks:
\textbf{(1) Traditional CBMs assume independence among concepts, overlooking essential inter-concept relationships that are instead inherent in medical imaging and critical for holistic reasoning. }
For example, lesion morphology interpretation depends on surrounding textures, vascular patterns, and anatomical structures.
\textbf{(2) CBMs typically require resource-intensive and time-consuming concept-level annotations to achieve satisfactory interpretability and maintain decent diagnosis results}. 
Expensive annotation costs unfortunately impede the scalability and application of CBMs in clinical scenarios.
To ease concept annotation, Semi-Supervised Concept Bottleneck Model (SSCBM) has been recently proposed \cite{sscbm}, leveraging unlabeled data for training. 
While promising, SSCBM relies on pseudo-labels derived from ImageNet pre-trained features, a strategy that introduces a significant domain gap when applied to clinical imaging. As a result, pseudo-labels lack medical fidelity, undermining both interpretability and downstream diagnostic accuracy~\cite{liu2023reducing,li2020domain}. 
\textit{\textbf{Addressing this requires a new framework that not only models the semantic dependencies between concepts, but also generates pseudo-labels grounded in domain-specific image semantics, which we tackle in this work.}}

In this paper, we propose a novel Hypergraph-driven semi-supervised concept bottleneck framework, dubbed \textbf{\textit{HyperCBM}}, tailored for label-efficient, interpretable medical image diagnosis. 
As shown in Fig.~\ref{teaser}, HyperCBM is significantly superior to traditional methods like CEM~\cite{cem} in detecting clinical concepts and diagnosing under limited concept labels.
Our framework tackles the dual challenges of high-order inter-concept relationship modeling and domain-specific pseudo-label generation, \ 
thereby enhancing both the interpretability and performance of concept bottleneck models.

Our contributions are summarized as: 
\begin{itemize}
    \item A semi-supervised concept bottleneck framework, which is the first design for medical imaging, improving both label efficiency and interpretability.
    \item A hypergraph-enhanced concept representation learning (HECRL) introduced to model high-order inter-concept relationships, enhancing diagnostic reasoning accuracy.
    \item A Hypergraph Image Dynamic Pseudo-labeling (HIDP) generation strategy developed, leveraging adaptive features to robustly exploit unlabeled data.
    \item Extensive experiments are conducted on a newly curated Placenta Accreta Spectrum dataset, public breast ultrasound and dermoscopic image datasets, demonstrating its effectiveness and state-of-the-art performance against existing methods.
\end{itemize}

\begin{figure*}[t]
    \centering
    \includegraphics[width=0.95\textwidth]{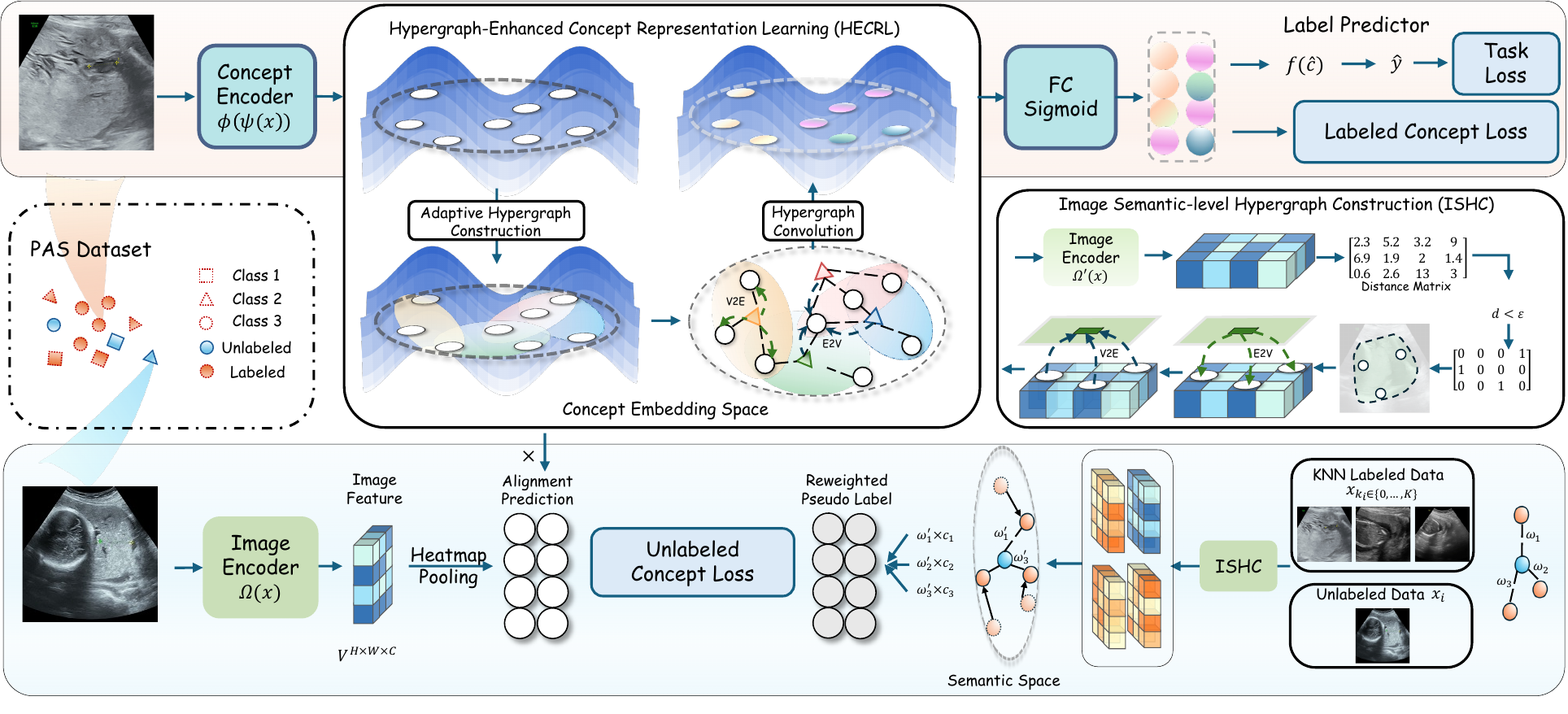}
    \vspace{-2mm}
    \caption{\textbf{Overview of HyperCBM}, a hypergraph-driven semi-supervised concept bottleneck model for ultrasound imaging. The framework integrates Hypergraph-Enhanced Concept Representation Learning (HECRL) for high-order inter-concept modeling via adaptive hypergraph propagation, and Hypergraph Image Dynamic Pseudo-labeling (HIDP) for reliable pseudo-label generation.}\label{framework}
    \vspace{-4mm}
\end{figure*}

\input{2-related}

\section{Method}
In this section, we introduce our method \textbf{HyperCBM} enabling explainable and data-efficient medical image diagnosis. \
We first present its preliminaries. Then, we provide a detailed description of the \textbf{Hypergraph-Enhanced Concept Representation Learning} (HECRL) and \textbf{Hypergraph Image Dynamic Pseudo-labeling} (HIDP) components. 
The overall framework is illustrated in Fig.~\ref{framework}. 

\subsection{Preliminary: CBM and SSCBM}
CBM~\cite{cbm} is a class of interpretable models designed to predict a target \( y \in Y \) from an input \( x \in X \) through an intermediate concept space \( C \). 
    The concept set \( C = \{p_1, \dots, p_t\} \) consists of \( t \) binary concepts provided by experts. 
The training dataset is represented as \( D = \{(x(i), y(i), c(i))\}_{i=1}^N \), where for each sample 
\( i \), \( x(i) \in X \subseteq \mathbb{R}^d \) is the input image, 
\( y(i) \in Y \subseteq \mathbb{R}^l \) denotes the label with \( l \) classes, and \( {c}^{(i)} =(c_i^1, \cdots, c_i^t) \in \{0, 1\}^t \) is 
the concept vector indicating the presence or absence of each concept. 
CBM learns two mappings: 
a \textbf{concept encoder} \( g: \mathbb{R}^d \rightarrow \mathbb{R}^t \), which transforms the input \( x \) into 
the concept space \( \hat{c} = g(x) \), and a \textbf{label predictor} \( f: \mathbb{R}^t \rightarrow \mathbb{R}^l \), which 
maps the concept vector \( \hat{c} \) to the final prediction \( \hat{y} = f(\hat{c}) \).

CEM~\cite{cem} mitigates CBM-induced performance degradation using high-dimensional concept embeddings. \
For each input \( x \), CEM generates \( t \) concept embeddings \( \hat{c}_1, \hat{c}_2, \dots, \hat{c}_t \), \
where each concept \( \hat{c}_i \) is represented by two embeddings \( \hat{c}^+_i, \hat{c}^-_i \in \mathbb{R}^m \), corresponding to the ``TRUE'' and ``FALSE'' states of the concept respectively. These embeddings are generated \ 
using a DNN \( \psi(x) \) to produce a latent representation \( h \in \mathbb{R}^{n} \), followed by concept-specific fully connected layers:
\(
\hat{c}_i = \phi_i(h).
\)
A differential scoring function \( s: \mathbb{R}^{2m} \rightarrow [0, 1] \) is used to align the embeddings with ground truth concepts, predicting the probability \( \hat{p}_i = \sigma(W_s [\hat{c}^+_i, \hat{c}^-_i]^\top + b_s) \) \ 
of concept \( c_i \) being active. The final concept embedding is computed as:
\begin{equation*}
    \bm{\hat{c}}_{i}:=\hat{p}_i\hat{c}_{i}^{+}+(1-\hat{p}_i)\hat{c}_{i}^{-}
\end{equation*}

CEM generates high-quality concept embeddings enriched with semantic information, enhancing both interpretability and task accuracy. However, this benefit comes at the cost of dense expert annotations on concepts.
SSCBM~\cite{sscbm} extends CEM by leveraging limited labeled data with abundant unlabeled data, reducing the concept annotation burden.
In SSCBM, the input set $\mathcal{X}$ is divided into two disjoint subsets for semi-supervised setting:
\(
\mathcal{X} = \mathcal{X}_{L} \cup \mathcal{X}_{U},
\)
where $\mathcal{X}_{L}$ represents a small subset of labeled data and $\mathcal{X}_{U}$ denotes the remaining unlabeled data, \ 
with $|\mathcal{X}_{L}| \ll |\mathcal{X}_{U}|$. For $x^{(j)} \in \mathcal{X}_{L}$, both concept annotations $c^{(j)}$ and class \ 
labels $y^{(j)}$ are available. For $x^{(i)} \in \mathcal{X}_{U}$, only the class label $y^{(i)}$ is accessible. \
Under these settings, given the combined training dataset $\mathcal{D} = \mathcal{D}_{L} \cup \mathcal{D}_{U}$, \ 
where $\mathcal{D}_{L} = \{(x^{(j)}, y^{(j)}, c^{(j)})\}_{j=1}^{|\mathcal{X}_{L}|}$ and $\mathcal{D}_{U} = \{(x^{(i)}, y^{(i)})\}_{i=1}^{|\mathcal{X}_{U}|}$, \ 
the objective is to jointly leverage both labeled and unlabeled data to train a CEM.

\subsection{Hypergraph-Enhanced Concept Representation Learning}
\label{sec:hypergraph_model}
Traditional CBMs overlook high-order semantic relations between concepts, which are crucial for capturing complex patterns in medical imaging, while the hypergraph structure effectively models multi-way correlations, representing their dependencies beyond the pairwise connections of traditional graphs.
Specifically, we construct an image-level concept embedding hypergraph $\mathcal{H}_c = (\mathcal{V}_c, \mathcal{E}_c, \mathcal{W}_c)$ for each image in a batch, where $\mathcal{V}_c$ denotes $t$ concept embeddings in total $\{\bm{\hat{c}}^i\}_{i=1}^t$, with each $\bm{\hat{c}}^i \in \mathbb{R}^{m}$, in which $m$ is the dimensionality of a single concept embedding. $\mathcal{H}_c$ is instantiated once per image, while the subsequent HGNN+ layers share weights across all images in a batch.
Here, $\mathcal{E}_c$ represents adaptively formed concept clusters, while $\mathcal{W}_c$ reflects the relative importance of each hyperedge. The detailed process is described below.

\subsubsection{Adaptive Hypergraph Construction}
Considering high-order semantic relationships, we first quantify pairwise relationships between concepts using cosine similarity in each image. 
This measure ensures that hyperedges are formed among semantically related concepts:
$ S_{ij} = \frac{\bm{\hat{c}}^i \cdot \bm{\hat{c}}^j}{\|\bm{\hat{c}}^i\|\|\bm{\hat{c}}^j\|} $.
\noindent The resulting similarity matrix $S \in \mathbb{R}^{t \times t}$ serves as the foundation for adaptive hyperedge construction.
Traditional methods often rely on fixed $k$-nearest neighbors, which may not reflect the varying semantic density among concepts. \ 
To address this, we define an adaptive target neighborhood size:
\begin{equation}
    k_{\text{init}} = \max\left(\lfloor t \times \text{initial\_ratio} \rfloor, k_\text{min}\right), 
\end{equation}
\noindent where $\text{initial\_ratio}$ and $k_\text{min}$ are predefined hyperparameters controlling the neighborhood scale. The outer $\max(\cdot)$ selects the densest local region as an upper bound so that even concepts located in tight clusters retain sufficient neighbours.

To ensure consistent semantic cohesion, we compute a global similarity threshold $\tau$ based on the average similarity across the top-$k_{\text{init}}$ neighbors:
\begin{equation}
    \tau = \frac{1}{t}\sum_{i=1}^t \frac{1}{k_{\text{init}}}\sum_{j \in \text{top-}k_{\text{init}}(S_{ij})} S_{ij}.
\end{equation}
Based on $\tau$, we update the adaptive neighborhood size $\hat{k}$ and construct hyperedges for each concept:
\begin{align}
    \hat{k} &= \max\left(\min\left(\max_i \sum_j \mathds{1}_{S_{ij} > \tau}, t-1\right), k_\text{min}\right), \\
    e^i_c &= \{ c^j \mid S_{ij} > \tau \text{ and among top-}\hat{k} \text{ similarities} \}.
\end{align}
\noindent Each such set \( e_i^c \subset \mathcal{V}_c \) forms a hyperedge centered on concept \( c^i \), capturing its most semantically related neighbors under a similarity threshold. Collectively, these egocentric hyperedges define the full hyperedge set \( \mathcal{E}_c = \{e^1_c, e^2_c, \dots, e^t_c\} \), which constitutes the concept-level hypergraph \( \mathcal{H}_c = (\mathcal{V}_c, \mathcal{E}_c, \mathcal{W}_c) \). 
This adaptive process ensures that hyperedges capture semantically coherent concept clusters, providing a flexible and robust representation of high-order dependencies.

\subsubsection{Attention-driven Hyperedge Weighting}
To ensure that the model prioritizes clinically relevant concept clusters, we introduce an \textit{attention-driven 
weighting mechanism}. This mechanism dynamically assigns weight scores to hyperedges, highlighting semantically salient 
relationships.

For each hyperedge $e\!\in\!\mathcal{E}_c$ that connects $\hat{k}$ concepts, we aggregate the corresponding concept embeddings $C_c^e \in \mathbb{R}^{\hat{k} \times m}$ and project the embeddings into a shared $d_a$‑dimensional latent space:
\begin{equation}
    Q = W_Q C_c^e, K = W_K C_c^e, V = W_V C_c^e,
\end{equation}
where $W_Q,W_K,W_V\in\mathbb{R}^{m\times d_a}$ are learnable parameters and $Q,K,V\in\mathbb{R}^{\hat{k}\times d_a}$.
We compute attention scores $\alpha^e_c$ to capture the importance of each hyperedge by its internal semantic consistency: $\alpha^e_c = \text{softmax}\left(QK^\top/\sqrt{d}\right) \in \mathbb{R}^{\hat{k}\times\hat{k}}$. 
$\alpha^e_c $ reflects the alignment of concept embeddings within a hyperedge. 
We then derive a scalar, unnormalised importance by aggregating the attention-reweighted value features across nodes within the hyperedge.
Specifically, we compute the mean of attention-weighted value features $ V^e_c \in \mathbb{R}^{\hat{k} \times d_a}$ to obtain a representative feature vector $ \bar{V}^e_c \in \mathbb{R}^{d_a}$ that captures the collective semantic information. 
\begin{equation}
  \bar{V}^{e}_{c}
  \;=\;
  \frac{1}{\hat{k}}
  \sum_{i=1}^{\hat{k}}
  \bigl(\alpha^{e}_{c}V\bigr)_{i,:}, \quad
    w^{e}_{c}
  \;=\;
  \bigl\lVert \bar{V}^{e}_{c} \bigr\rVert_{2}
\end{equation}
The $\ell_2$-norm then quantifies the semantic coherence and feature activation intensity of this aggregated representation, serving as a measure of the hyperedge's clinical relevance and conceptual consistency.
We then apply a softmax operation over all hyperedges belonging to the same image. 
This attention-driven weighting mechanism ensures that the model dynamically focuses on concept clusters that are more relevant to clinical outcomes:
\begin{equation}
\tilde{w}_c^{e} \;=\;
\frac{\exp\!\bigl(w_c^{e}\bigr)}
     {\displaystyle\sum_{e'\in\mathcal{E}_c}\exp\!\bigl(w_c^{e'}\bigr)} .
\end{equation}
The normalised coefficients $\{\tilde{w}_c^{e}\}_{e\in\mathcal{E}_c}$ populate
the diagonal matrix $\mathcal{W}_c$, which is used as the hyperedge weight matrix
in subsequent HGNN$^{+}$ propagation.

After constructing the hypergraph with adaptively weighted hyperedges, we employ \textit{HGNN+ layers}~\cite{hgnn+} for high-order semantic reasoning. 
Unlike traditional HGNN \cite{hgnn}, HGNN+ integrates vertex-to-hyperedge (V2E) aggregation and hyperedge-to-vertex (E2V) propagation 
into a single compact formulation. 
Given the concept embeddings $\bm{\hat{c}_i}^{(l)}$ at the $l$-th layer, the HGNNConv+ layer is defined as:
\begin{equation}
    \bm{\hat{c}_i}^{(l+1)} = \sigma\left(D_v^{-1} H \mathcal{W}_c D_e^{-1} H^\top \bm{\hat{c}_i}^{(l)} \Theta^{(l)}\right) + \bm{\hat{c}_i}^{(l)},
\end{equation}
where $H$ is the hypergraph incidence matrix encoding V2E relations, $D_v$ and $D_e$ are the vertex and hyperedge degree matrices for E2V propagation, $\Theta^{(l)}$ represents the learnable parameters at layer $l$, and $\sigma(\cdot)$ is a non-linear activation function. The final layer result can be denoted as $\bm{\hat{c}_{\text{concept}}}$.
Note that residual connections are applied to enhance training stability and mitigate over-smoothing. The resulting $\bm{\hat{c}}_{\text{concept}}$ is forwarded to unlabelled data concept prediction and subsequently to the final classification head.

\subsection{Hypergraph Image Dynamic Pseudo-labeling}

To overcome the limitations of pseudo-labeling in SSCBM \cite{sscbm}, particularly the lack of domain-specific semantics, we propose Hypergraph Image Dynamic Pseudo-labeling (HIDP). HIDP constructs an image-level hypergraph on semantic feature maps to improve pseudo-label selection through hypergraph-based feature aggregation, dynamic pseudo-label generation, and enhanced alignment.

\subsubsection{Hypergraph Construction on Semantic Feature Maps}
Given an unlabeled image sample $x$, we extract its semantic image feature map as a vertex using an image encoder. To incorporate contextual relationships, we select the $K$-nearest labeled samples based on similarity in the feature space. 
To capture high-order spatial correlations among these feature maps, we construct an image-level hypergraph $\mathcal{H} = (\mathcal{V}, \mathcal{E})$. \ 
Following a distance-based hypergraph paradigm \cite{hyperyolo}, each spatial feature point $(p,q)$ in the $h \times w$ grid corresponds to a vertex in the hypergraph. \ 
For each vertex $v$, an $\epsilon$-ball hyperedge includes all neighboring vertices $u$ satisfying a distance constraint in the feature space. \ 
This construction ensures that each hyperedge represents a local semantic neighborhood within the image feature space. Note that hyperedges are confined to a single image; cross-image information is introduced only through the subsequent pseudo-label aggregation step.

\vspace{-2mm}
\subsubsection{Hypergraph-driven Dynamic Pseudo-label Generation}
To model high-order dependencies among feature points, we apply a single hypergraph convolutional layer with residual connections as:
\begin{equation}
    \hat{V} = V + D_v^{-1} H \mathcal{W} D_e^{-1} H^\top V \Theta,
\end{equation}
where \(V\) represents vertex features, \(H\) the hypergraph incidence matrix, \(D_v\) and \(D_e\) the vertex and hyperedge degree matrices, respectively, and \(\Theta\) the learnable parameters. 
This operation enables each vertex to aggregate information from its connected hyperedges, capturing contextual semantic relationships within the image.
Then we compute the Euclidean distance \(d_i = \| \hat{V}^u - \hat{V}^l_i \|\) between the feature of the unlabeled sample \(\hat{V}^u\) and its \(K\) neighbors \(\{\hat{V}^l_i\}_{i=1}^K\) with labeled concepts \(\{c^l_i\}_{i=1}^K\). 
The final pseudo-label \(\hat{c}_{\text{pseudo}}\) is calculated as:
\begin{equation}
    \omega_i = \frac{\left( 1/d_i \right)}{\sum_{j=1}^{K} \left( 1/d_j \right)}, 
    \bm{\hat{c}}_{\text{pseudo}} = \sum_{i=1}^K \omega_i \cdot c^l_i.
\end{equation}
\subsubsection{Semi-supervised Training}
To enhance concept interpretability while maintaining classification performance, for labeled data, we define concept loss $\mathcal{L}_{c}$ using binary cross-entropy (BCE) to enforce consistency between predicted concepts $\bm{\hat{c}}$ and ground-truth labels $\bm{c}$. 
For unlabeled data, we introduce the alignment loss $\mathcal{L}_{align}$ to enforce coherence between concept embeddings and image features. We derive $\bm{\hat{c}}_{\text{hyper}}$ from the semantic feature map $V$ and updated concept embeddings $\bm{\hat{c}}_\text{concept}$ via heatmap-based operations and align it with similarity-based concept labels $\bm{\hat{c}}_{\text{pseudo}}$ using BCE as $\mathcal{L}_{align}$. 
The task loss $\mathcal{L}_{task}$ ensures accurate classification by mapping concept embeddings $\hat{\boldsymbol{c}}_\text{concept}$ to final predictions $\hat{\boldsymbol{y}}$ via a label predictor $g(\cdot)$, optimized with categorical cross-entropy. 
The overall objective is formulated as 
\begin{equation}
    \mathcal{L} = \mathcal{L}_{task} + \lambda_1 \mathcal{L}_{c} + \lambda_2 \mathcal{L}_{align},
\end{equation}
where $\lambda_1$ and $\lambda_2$ balance interpretability and classification accuracy in semi-supervised setting (\textit{See Implementation Details}).
We define \( \mathcal{L}_c = \text{BCE}(\hat{c}, c) \) for labeled samples and \( \mathcal{L}_{\text{align}} = \text{BCE}(\hat{c}_{\text{concept}}, \hat{c}_{\text{pseudo}}) \) for unlabeled ones, where \( \hat{c}_{\text{concept}} \) is obtained via attention-based decoding.

\begin{table*}[t]
\centering
\caption{Results of concept and task accuracy and AUC across PAS, BrEaST, and SkinCon datasets with different labeled data ratios. `*' denotes the model trained in a fully supervised setting. }
\resizebox{\linewidth}{!}{%
\begin{tabular}{cc|cccc|cccc|cccc}
\toprule[0.15em]
\multirow{2}{*}{\textbf{Method}} & \multirow{2}{*}{\textbf{Labeled Ratio}} &
\multicolumn{4}{c|}{\textit{PAS}} &
\multicolumn{4}{c|}{\textit{BrEaST}} &
\multicolumn{4}{c}{\textit{SkinCon}} \\
\cmidrule(lr){3-6} \cmidrule(lr){7-10} \cmidrule(lr){11-14}
& & Concept Acc. & Class Acc. & Concept AUC & Class AUC
  & Concept Acc. & Class Acc. & Concept AUC & Class AUC
  & Concept Acc. & Class Acc. & Concept AUC & Class AUC \\
\midrule
\multirow{7}{*}{\textbf{HyperCBM}}
& 0.01 & 70.21$\pm$1.44 & 57.33$\pm$9.88 & 52.67$\pm$1.97 & 76.21$\pm$7.48 
        & 64.26$\pm$6.43 & 65.88$\pm$4.57 & 53.75$\pm$2.06 & 67.32$\pm$14.47 
        & 86.08$\pm$1.06 & 72.33$\pm$1.11 & 50.52$\pm$0.68 & 57.57$\pm$2.54 \\
& 0.1  & 81.61$\pm$0.88 & 76.89$\pm$3.85 & 64.43$\pm$1.27 & 88.40$\pm$2.51 
        & 72.49$\pm$3.90 & 65.49$\pm$7.08 & 56.66$\pm$3.96 & 70.31$\pm$9.70 
        & 88.34$\pm$0.51 & 73.99$\pm$0.96 & 54.59$\pm$2.03 & 70.53$\pm$3.09 \\
& 0.2  & 83.07$\pm$0.70 & 76.59$\pm$2.84 & 67.08$\pm$2.05 & 90.57$\pm$1.60 
        & 71.99$\pm$2.49 & 66.27$\pm$9.06 & 55.26$\pm$3.51 & 72.00$\pm$8.31 
        & 89.06$\pm$0.25 & 75.76$\pm$0.99 & 56.71$\pm$1.54 & 74.39$\pm$1.68 \\
& 0.4  & 84.19$\pm$1.11 & 78.82$\pm$4.18 & 68.20$\pm$3.00 & 90.48$\pm$3.03 
        & 76.25$\pm$2.81 & 75.29$\pm$4.22 & 61.93$\pm$4.69 & 80.11$\pm$2.42 
        & 89.87$\pm$0.22 & 75.86$\pm$0.80 & 60.14$\pm$1.23 & 75.34$\pm$0.98 \\
& 0.6  & 85.57$\pm$0.83 & 80.15$\pm$1.65 & 71.25$\pm$1.91 & 93.33$\pm$2.46 
        & 77.81$\pm$1.40 & 73.33$\pm$7.40 & 62.70$\pm$4.70 & 80.75$\pm$5.15 
        & 90.13$\pm$0.45 & 75.48$\pm$0.29 & 61.03$\pm$0.89 & 77.64$\pm$1.77 \\
& 0.8  & 85.57$\pm$0.86 & 81.93$\pm$3.49 & 71.89$\pm$1.36 & 91.95$\pm$1.44 
        & 76.08$\pm$1.83 & 64.70$\pm$9.92 & 57.52$\pm$4.85 & 74.49$\pm$7.60 
        & 90.12$\pm$0.34 & 75.92$\pm$0.93 & 60.73$\pm$1.86 & 75.30$\pm$1.14 \\
\midrule
\rowcolor[HTML]{DBFFDB}\textbf{HyperCBM*} & 1.0  & \textbf{84.93$\pm$0.93} & \textbf{79.85$\pm$2.36} & \textbf{71.45$\pm$1.90} & \textbf{92.51$\pm$2.18} 
        & \textbf{78.88$\pm$1.33} & \textbf{74.12$\pm$7.78} & \textbf{64.13$\pm$3.34} & 80.61$\pm$6.82 
        & \textbf{90.43$\pm$0.22} & 77.48$\pm$1.20 & \textbf{62.34$\pm$0.61} & \textbf{79.86$\pm$1.45} \\
CBM* &   1.0      & 79.85$\pm$2.50 & 75.85$\pm$10.51 & 64.64$\pm$7.29 & 88.91$\pm$8.34 
                   & 76.30$\pm$0.29 & 70.98$\pm$10.03 & 59.08$\pm$2.55 & 79.88$\pm$4.65 
                   & 90.24$\pm$0.32 & 72.43$\pm$1.21 & 61.72$\pm$0.83 & 69.37$\pm$1.73 \\
CEM* &  1.0       & 78.14$\pm$2.24 & 77.78$\pm$2.52 & 61.67$\pm$2.66 & 91.09$\pm$1.52 
                   & 76.36$\pm$3.45 & 71.76$\pm$6.40 & 58.92$\pm$5.81 & \textbf{81.45$\pm$2.58} 
                   & 90.33$\pm$0.33 & 76.10$\pm$2.02 & 62.22$\pm$0.82 & 78.42$\pm$1.33 \\
ResNet* &  --    & N/A           & 78.22$\pm$2.55 & N/A            & 91.09$\pm$2.09 
                   & N/A           & 74.12$\pm$3.37 & N/A            & 80.50$\pm$3.85 
                   & N/A           & \textbf{77.88$\pm$0.92} & N/A            & 77.55$\pm$0.80 \\
\bottomrule[0.15em]
\end{tabular}%
}
\label{tab:ratio}
\end{table*}

\begin{table*}[t]
    \centering
    \caption{Evaluation with concept labeled ratios of 0.1, 0.4 on three datasets. Best results are in \textbf{bold}, second-best are \underline{underlined}.}
    \resizebox{\linewidth}{!}{%
    \begin{tabular}{cc|cccc|cccc|cccc}
        \toprule[0.15em]
        \multirow{2}{*}{\textbf{Labeled Ratio}} & \multirow{2}{*}{\textbf{Method}} 
        & \multicolumn{4}{c|}{\textit{PAS}} 
        & \multicolumn{4}{c|}{\textit{BrEaST}} 
        & \multicolumn{4}{c}{\textit{SkinCon}} \\
        \cmidrule(lr){3-6} \cmidrule(lr){7-10} \cmidrule(lr){11-14}
        & & Concept Acc. & Class Acc. & Concept AUC & Class AUC 
          & Concept Acc. & Class Acc. & Concept AUC & Class AUC 
          & Concept Acc. & Class Acc. & Concept AUC & Class AUC \\
        \midrule
        \multirow{4}{*}{0.1}
        & CBM    & \underline{77.54$\pm$1.77} & 65.93$\pm$11.52 & \underline{57.63$\pm$5.05} & 83.14$\pm$9.55 
                 & 68.16$\pm$0.96 & \underline{61.96$\pm$5.63} & 52.67$\pm$1.97 & 47.02$\pm$12.94 
                 & \underline{88.17$\pm$0.44} & 72.43$\pm$1.21 & 50.48$\pm$0.11 & 53.70$\pm$3.68 \\
        & CEM    & 71.23$\pm$1.82 & 69.92$\pm$3.88 & 53.52$\pm$2.82 & 86.03$\pm$3.06 
                 & 68.13$\pm$2.28 & 61.57$\pm$5.35 & 50.79$\pm$2.51 & 58.08$\pm$7.19 
                 & 87.94$\pm$0.94 & \textbf{74.14$\pm$0.14} & \underline{53.54$\pm$2.81} & \textbf{70.69$\pm$3.54} \\
        & SSCBM  & 73.77$\pm$2.55 & \underline{73.78$\pm$2.99} & 56.28$\pm$4.04 & \textbf{88.68$\pm$1.85} 
                 & \underline{71.43$\pm$3.09} & 61.57$\pm$5.77 & \underline{52.90$\pm$4.26} & \underline{62.87$\pm$10.62} 
                 & 87.97$\pm$0.99 & 73.92$\pm$0.63 & 53.08$\pm$2.47 & 69.93$\pm$3.59 \\
   \rowcolor[HTML]{DBFFDB}     &  \textbf{Ours} 
                 & \textbf{81.61$\pm$0.88} & \textbf{76.89$\pm$3.85} & \textbf{64.43$\pm$1.27} & \underline{88.40$\pm$2.51} 
                 & \textbf{72.49$\pm$3.90} & \textbf{65.49$\pm$7.08} & \textbf{56.66$\pm$3.96} & \textbf{70.31$\pm$9.70} 
                 & \textbf{88.34$\pm$0.51} & \underline{73.99$\pm$0.96} & \textbf{54.59$\pm$2.03} & \underline{70.53$\pm$3.09} \\
        \midrule
        \multirow{4}{*}{0.4}
        & CBM    & \underline{79.33$\pm$2.14} & 73.63$\pm$9.14 & \underline{62.72$\pm$6.01} & 87.66$\pm$8.81 
                 & 70.93$\pm$1.07 & 61.57$\pm$5.35 & 52.28$\pm$2.01 & 50.86$\pm$7.83 
                 & \underline{89.79$\pm$0.18} & 72.43$\pm$1.21 & 54.93$\pm$1.77 & 64.25$\pm$4.13 \\
        & CEM    & 70.13$\pm$1.60 & 70.37$\pm$3.71 & 51.47$\pm$1.94 & 85.16$\pm$3.53 
                 & 73.50$\pm$1.78 & 66.28$\pm$8.54 & 54.97$\pm$2.68 & \underline{80.45$\pm$4.49} 
                 & 88.82$\pm$1.22 & 74.49$\pm$1.11 & \underline{57.42$\pm$3.19} & \underline{72.69$\pm$3.66} \\
        & SSCBM  & 78.16$\pm$3.81 & \underline{76.44$\pm$1.65} & 60.51$\pm$3.67 & \underline{89.96$\pm$2.02} 
                 & \underline{73.89$\pm$2.06} & \underline{72.94$\pm$6.83} & \underline{56.17$\pm$3.23} & \textbf{81.46$\pm$2.18} 
                 & 88.78$\pm$1.31 & \underline{74.89$\pm$0.76} & 56.46$\pm$2.83 & 72.16$\pm$3.63 \\
     \rowcolor[HTML]{DBFFDB}   &  \textbf{Ours} 
                 & \textbf{84.11$\pm$1.11} & \textbf{78.67$\pm$4.18} & \textbf{68.73$\pm$3.00} & \textbf{92.41$\pm$5.91} 
                 & \textbf{76.25$\pm$2.81} & \textbf{75.29$\pm$4.22} & \textbf{61.93$\pm$4.69} & 80.11$\pm$2.42 
                 & \textbf{89.87$\pm$0.22} & \textbf{75.86$\pm$0.80} & \textbf{60.14$\pm$1.23} & \textbf{75.34$\pm$0.98} \\
        \bottomrule[0.15em]
\end{tabular}
}
\label{tab:perform}
\end{table*}

\section{Experiments}
\subsection{Datasets and settings}
\label{sec:imple_detail}
We evaluate our framework on three datasets: a newly collected ultrasound Placenta Accreta Spectrum grading dataset (\textit{\textbf{PAS}}), a public ultrasound breast lesion diagnosis dataset \textit{\textbf{BrEaST}}~\cite{breast}, and a public dermoscopic imaging dataset \textit{\textbf{Skincon}}~\cite{skincon1}.

\textbf{PAS} is a newly collected placenta-accreta-spectrum ultrasound dataset that contains 671 scans acquired from multiple vendors and annotated with three severity levels and 45 clinically curated concepts. Candidate concepts are extracted by HuatuoGPT-Vision and confirmed by two senior obstetric radiologists. 
\textbf{BrEaST} originally contains 256 breast ultrasound scans with 7 concepts from BI-RADS descriptors and 3 diagnostic labels. Following \cite{chenhao}, we exclude the Normal category, and finally use the 254 abnormal images with malignant and benign categories. 
\textbf{SkinCon} selects 3,205 dermoscopic images from Fitzpatrick-17k \cite{fitz17k}, covering Malignant, Benign, and Non-neoplastic lesions. Two dermatologists densely labelled 48 clinical concepts, of which the 22 occurring in at least 50 images are retained, consistent with \cite{chenhao}.
For all three datasets, images are centre-cropped and resized to \(224\times224\); data are randomly split into training, validation, and test subsets in a 7 : 1 : 2 ratio. Every experiment is repeated with five fixed random seeds, and results are reported as mean ± standard deviation. 

We compare with the following baselines: CBM~\cite{cbm}, CEM~\cite{cem}, and SSCBM~\cite{sscbm} in full-supervised and semi-supervised settings.
For evaluation, we use the Area Under the Receiver Operating Characteristic Curve (AUC), Accuracy (ACC) as disease diagnosis and concept detection tasks. 

\noindent\textbf{Implementation Details. }
All experiments were conducted on a single NVIDIA RTX 4090 GPU. 
We employ a ResNet~\cite{he2016deep} backbone as the image encoder, specifically using ResNet34 for the \textit{BrEaST} and \textit{PAS} datasets, and ResNet50 for the \textit{SkinCon} dataset. 
Both the concept adapters and the aggregator are implemented as fully-connected layers. 
The image encoder within the concept encoder, denoted as $ \psi(x) $, and the unlabeled image encoder $\Omega(x)$ share their weights. To strictly enforce the stability of pseudo-label generation during training, the image encoder $\Omega^{'}(x)$ utilized to extract features for ISHC module remains frozen throughout each training epoch. Then the weights are updated at the end of each training epoch by synchronizing the latest optimized parameters of $ \psi(x) $.
We train all models end-to-end utilizing the Adam optimizer. The models for \textit{BrEaST}, \textit{PAS}, and \textit{SkinCon} are trained for a maximum of 150, 250, and 100 epochs, respectively. The corresponding learning rates are set to $5\times10^{-4}$ for \textit{BrEaST} and \textit{PAS}, and $5\times10^{-3}$ for \textit{SkinCon}. To prevent overfitting, we employ an early stopping strategy with a patience of 5 epochs, monitoring the validation loss. 
For HECRL, the number of nearest neighbors for a hyperedge, $k_\text{min}$, is set to 2 for \textit{BrEaST} and 3 for the more concept-diverse \textit{PAS} and \textit{SkinCon} datasets. The radius for the $\epsilon$-ball hyperedge, $\epsilon$ is fixed at 6 across all. The loss balancing weights ($\lambda_1$,$\lambda_2$) are empirically configured to (0.5,0.1), (1.0,0.1), and (2.0,1.0) for the \textit{BrEaST}, \textit{PAS}, and \textit{SkinCon} datasets, respectively.

\begin{figure}[t]
    \centering
\includegraphics[width=0.85\linewidth]{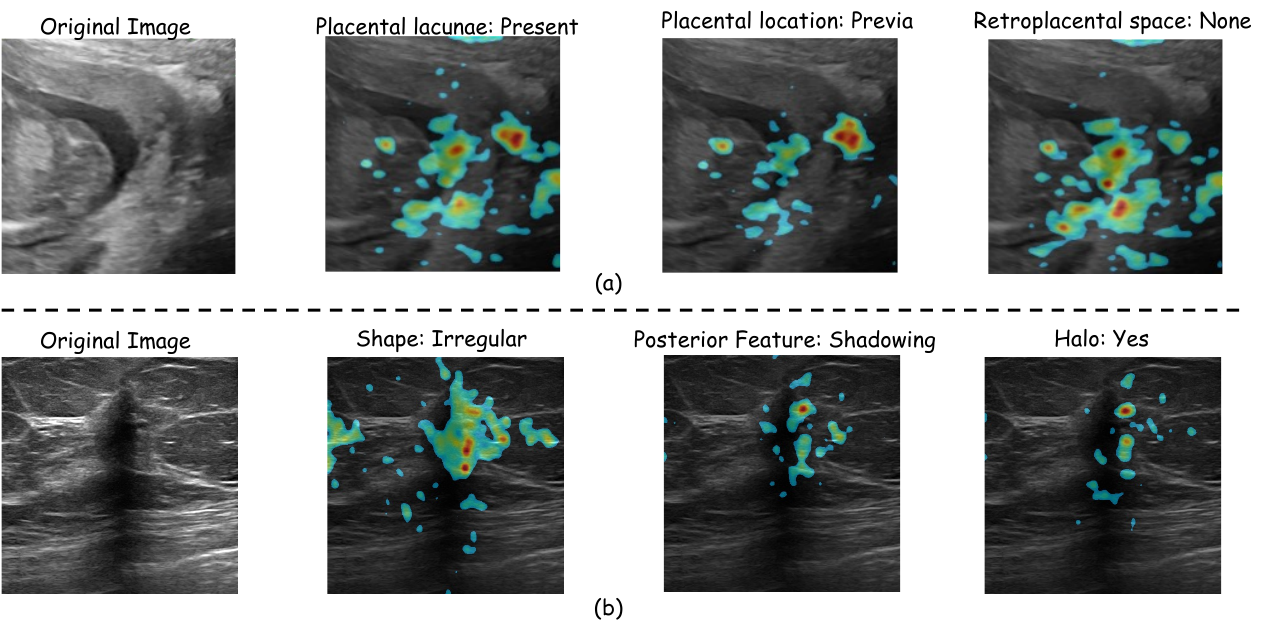}
\caption{\textbf{Interpretability Visualization}: (a) Concept saliency maps on the PAS dataset, highlighting learned concepts (\eg, placental lacunae, retroplacental space). (b) Concept saliency maps on the BrEaST dataset, capturing key diagnostic features (\eg, irregular shape, posterior features).}
\label{fig:heatmap} 
\end{figure}

\begin{figure}[t]
    \centering
\includegraphics[width=0.9\linewidth]{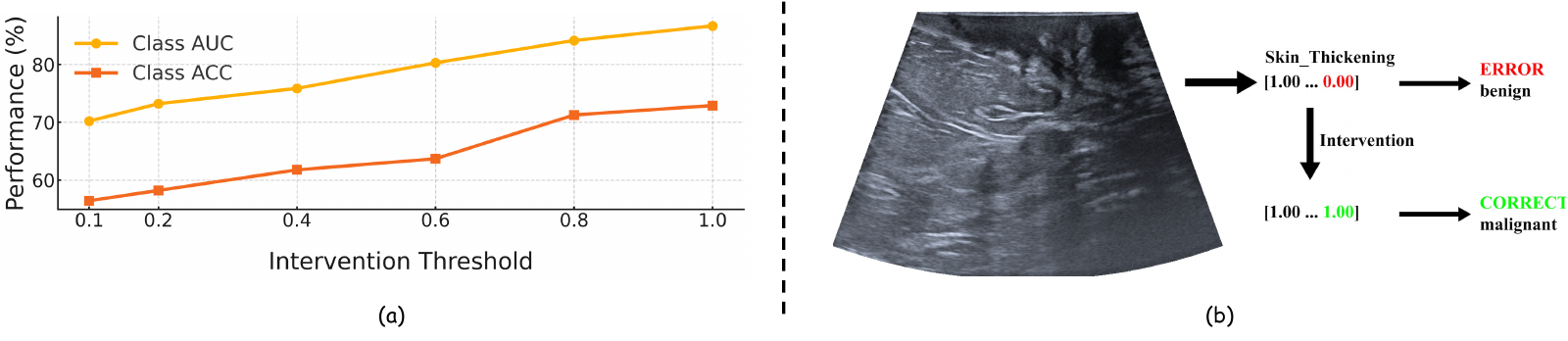}
\caption{\textbf{Test-time intervention results on the PAS dataset}: (a) Any concept whose score exceeds the intervention threshold is forced to zero. This intervention causes a nearly monotonic degradation in diagnosis.  (b) An example demonstration of test-time intervention, where correcting "Skin Thickening" shifts the prediction from benign to malignant, improving diagnosis results and demonstrating model applicability.}
\label{fig:intervention} 
\end{figure}

\subsection{Quantitative Comparison}
\textit{\textbf{In Tab.~\ref{tab:ratio}, we present a comprehensive evaluation of HyperCBM's performance in a semi-supervised setting, systematically varying the proportion of labeled data from 1\% to 80\%}} across three distinct datasets: PAS, BrEaST, and SkinCon. The results are benchmarked against several fully supervised models, including our own HyperCBM, traditional CBM, CEM, and a standard ResNet backbone.
On the PAS dataset, HyperCBM shows a clear and consistent improvement with increased supervision concept label ratio. Concept accuracy rises from 70.21\% to 85.57\%, and class accuracy increases from 57.33\% to 81.93\%. A key finding is the model's high data efficiency: with only 40\% of labeled data, HyperCBM's class accuracy (78.82\%) is already competitive with the fully supervised ResNet34 model (78.22\%) and superior to CBM (75.85\%) and CEM (77.78\%). 
The model's performance on the BrEaST and SkinCon datasets further validates its capabilities. On BrEaST, despite some performance fluctuations attributable to the dataset's limited number of concept annotations (7 concepts), HyperCBM with 40\% labeled data still surpasses other fully supervised concept models in task accuracy. On SkinCon, the performance progression is more stable, with our model at 20\% supervision (75.76\% class accuracy) nearing the performance of the fully supervised CEM (76.10\%).
Collectively, these experiments demonstrate that our semi-supervised HyperCBM is highly effective in low-ratio concept label regimes, achieving performance comparable or superior to fully supervised baselines with significantly less data. Thus, HyperCBM offers a robust solution that maintains high task accuracy while providing model interpretability, which is valuable in label-constrained scenarios.

\textit{\textbf{Table~\ref{tab:perform} ablates HyperCBM against leading semi-supervised baselines to verify its architectural advantages.}} The results yield two central conclusions: 1) HyperCBM consistently outperforms existing methods in low-data regimes, and 2) its performance advantage often scales with increased data availability. Even with only 10\% of labels, HyperCBM establishes a substantial performance margin. On the PAS dataset, it leads the next-best method by +3.11\% in class accuracy, while on BrEaST, it achieves top performance across all four metrics. 
Crucially, this performance gap widens at the 40\% label ratio, highlighting our model's superior scalability. For instance, its concept accuracy leads on PAS grows to +4.78\%, and it surpasses all competitors across all metrics on the SkinCon dataset.
The consistent outperformance and scalability strongly suggest that HyperCBM's architecture is fundamentally more effective at leveraging heterogeneous supervision signals. 
Unlike baselines whose gains may plateau, HyperCBM is more adept at integrating semi-supervised information, leading to more robust concept representations and superior task accuracy. 
This firmly establishes HyperCBM as a new state-of-the-art for semi-supervised concept-based learning.

\begin{table*}[t]
    \centering
    \caption{Ablation study on PAS and BrEaST datasets (labeled ratio = 0.1). Best results are in \textbf{bold}, second-best are \underline{underlined}.}
    \resizebox{0.95\linewidth}{!}{%
        \begin{tabular}{@{}cc|cccc|cccc@{}}
            \toprule[0.15em]
            \multirow{2}{*}{\textbf{HECRL}} & \multirow{2}{*}{\textbf{HIDP}} &
            \multicolumn{4}{c|}{\textit{PAS}} &
            \multicolumn{4}{c}{\textit{BrEaST}} \\
            \cmidrule(lr){3-6} \cmidrule(lr){7-10}
            & & Concept Acc. & Class Acc. & Concept AUC & Class AUC 
              & Concept Acc. & Class Acc. & Concept AUC & Class AUC \\
            \midrule
            - & -             & 73.64 ± 2.77 & 72.00 ± 2.75 & 55.82 ± 3.47 & 88.19 ± 1.25 
                              & 69.19 ± 3.52 & \textbf{65.49 ± 6.28} & 50.12 ± 4.42 & 69.09 ± 6.54 \\
            \checkmark & -     & 80.42 ± 1.88 & 72.44 ± 11.47 & 63.16 ± 4.37 & 87.33 ± 6.07 
                              & \underline{71.60 ± 3.12} & 63.53 ± 5.90 & \underline{55.41 ± 4.56} & \underline{70.11 ± 9.48}\\
            - & \checkmark     & \underline{80.64 ± 0.67} & \underline{75.70 ± 4.07} & \underline{63.86 ± 2.21} & \textbf{90.16 ± 1.39} 
                              & 70.65 ± 3.48 & 63.14 ± 3.38 & 54.11 ± 2.56 & 67.31 ± 6.04  \\
            \checkmark & \checkmark & \textbf{81.61 ± 0.88} & \textbf{76.89 ± 3.85} & \textbf{64.43 ± 1.27} & \underline{88.40 ± 2.51} 
                              & \textbf{72.49 ± 3.90} & \underline{65.49 ± 7.08} & \textbf{56.66 ± 3.96} & \textbf{70.31 ± 9.70} \\
            \bottomrule[0.15em]
        \end{tabular}%
    }
    \label{tab:ablation_study}
\end{table*}

\subsection{Interpretability and Test-time Intervention}
\noindent\textbf{Clinical Interpretability. } 
We generate concept activation maps with Grad-CAM~\cite{selvaraju2017grad} for PAS and BrEaST test images, rescaling the heat-map intensity by the corresponding concept scores. Two board-certified radiologists independently examined the overlays and agreed that the highlighted regions coincide with the intended medical concepts.
The upper-left panel in Fig.~\ref{fig:heatmap} shows the raw ultrasound image; the remaining panels depict selected concepts together with their activation maps (warmer colours indicate stronger evidence).  
Across both datasets, the highlighted areas match the regions that clinicians rely on for diagnosis, underscoring the practical plausibility of the model’s concept-level explanations.

\vspace{-2mm}
\noindent\textbf{Test-time Intervention. } 
To assess faithfulness, we perform test-time interventions by applying a series of confidence thresholds $\tau\!\in\!\{0.1,0.2,0.4,0.6,0.8,1.0\}$ to the concept prediction scores: any concept whose score exceeds $\tau$ is forced to zero, following \cite{chenhao}. A smaller $\tau$ therefore eliminates a larger fraction of high-confidence concepts, enabling us to examine how sensitive the final diagnosis is to the removal of concept evidence.
As shown in Fig. \ref{fig:intervention}, intervention causes a nearly monotonic degradation in diagnosis: with the harshest setting ($\tau=0.1$), \textbf{Class AUC} drops from 86.67 \% to 70.21 \% and \textbf{Class ACC} from 72.89 \% to 56.44 \%. Performance steadily recovers as $\tau$ increases, reaching 84.14 \% AUC and 71.26 \% ACC at $\tau=0.8$, and returning to baseline when no concepts are masked ($\tau=1.0$).
The consistent decline confirms that the classifier relies strongly on the predicted concepts: the more evidence we suppress, the larger the performance loss.

\vspace{-2mm}
\subsection{Ablation Study}
As shown in Tab.~\ref{tab:ablation_study}, we evaluate the contributions of HECRL and HIDP individually and in combination.
The primary role of HECRL is to significantly enhance the model's concept-level understanding. Its inclusion leads to a dramatic concept accuracy boost of +6.78\% on PAS and +2.41\% on BrEaST. This demonstrates HECRL's remarkable effectiveness in refining concept representations by modeling their inter-dependencies.
Complementing this, HIDP is instrumental in improving downstream task performance. It delivers a substantial +3.70\% gain in class accuracy on PAS, showcasing its strength in effectively mapping the learned concepts to the final prediction. This highlights its crucial role in bridging concept and task learning.

\vspace{-2mm}
\section{Broader Impacts and Limitations} \label{broader}
\vspace{-2mm}

\noindent\textbf{Broader Impacts.}
This work contributes to a broader shift from opaque prediction-driven medical AI toward more transparent, knowledge-guided, and human-aligned diagnostic systems. By structuring model reasoning around clinically meaningful concepts, HyperCBM helps bridge the gap between statistical pattern recognition and clinician-interpretable evidence, which is essential for building trustworthy AI in high-stakes healthcare. Its label-efficient design suggests a more scalable path for interpretable models in domains where expert annotation is scarce, costly, or unevenly distributed.

\vspace{-2mm}
\noindent\textbf{Limitations.}
HyperCBM still has several limitations. First, its interpretability depends on a predefined clinical concept set, which may not fully capture all subtle diagnostic cues used by clinicians. Second, the newly collected PAS dataset remains limited in scale, and larger multi-center validation is needed to further assess generalization across institutions, scanners, and patient populations. 
Third, as a semi-supervised framework, HyperCBM may be affected by noisy pseudo concept labels, especially when unlabeled samples exhibit substantial domain shifts.

\vspace{-2mm}
\section{Conclusion}
In this work, we design a semi-supervised concept bottleneck model \textit{HyperCBM} for interpretable medical image diagnosis, addressing the limitations of inter-concept
dependency modeling and pseudo-label reliability. 
By introducing a concept-level hypergraph for structured reasoning and an image-level hypergraph for adaptive pseudo-labeling, our framework improves both interpretability and diagnostic accuracy. 
Experiments on our collected PAS dataset, breast ultrasound and dermoscopic image public datasets demonstrate superior interpretable performance over existing CBMs methods.
This presents a powerful baseline for concept label-efficient medical image analysis and facilitates clinical applications.

\bibliographystyle{nips}
\bibliography{main}



\end{document}

%% file: math_commands.tex
\usepackage{amsmath,amsfonts,bm}

\def\eqref#1{equation~\ref{#1}}

\def\1{\bm{1}}

\DeclareMathAlphabet{\mathsfit}{\encodingdefault}{\sfdefault}{m}{sl}
\SetMathAlphabet{\mathsfit}{bold}{\encodingdefault}{\sfdefault}{bx}{n}



%% file: 2-related.tex
\vspace{-2mm}
\section{Related Work}
\noindent\textbf{Concept Bottleneck Models (CBMs)} have emerged as a promising approach for enhancing interpretability in deep learning. 
CBMs introduce an intermediate concept layer between input and prediction, enabling models to make decisions based on human-interpretable concepts. Early works~\cite{cbm,cem} demonstrated CBMs could improve generalization and transparency, though they suffered from performance degradation compared to traditional black-box models and required expensive manual annotations. To address this, recent efforts~\cite{chauhan2023interactive,oikarinen2023label} proposed interactive CBMs that selectively annotate concepts and label-free CBMs that eliminate the need for labeled concept data. 
However, they heavily rely on large language models like GPTs, which have reliability issues~\cite{lai2023faithful}, and severely undermine their interpretability.
Studies like \cite{magister2021gcexplainer,barbiero2024relational} map graphs to concept spaces using clustering and human-in-the-loop strategies to improve transparency. 
CBMs work in the image field also include the works of \cite{havasi2022addressing,kim2023probabilistic,sheth2023auxiliary}.
Despite this progress, few approaches explore label efficiency, achieving scalable CBMs by semi-supervised learning in clinical applications.

\vspace{-2mm}
\noindent\textbf{Semi-supervised learning (SSL)} is widely used in medical image diagnosis where labeled data is scarce and expensive to obtain. Classical SSL strategies include pseudo-labeling~\cite{kamraoui2021popcorn, li2020self,wu2024semi,liu2025autoregressive}, consistency regularization~\cite{gu2025dual, xiao2025cuamt}, and hybrid methods like MixMatch~\cite{berthelot2019mixmatch} and FixMatch~\cite{sohn2020fixmatch}. In diagnosis-oriented tasks, POPCORN~\cite{kamraoui2021popcorn} introduces progressive pseudo-labeling guided by feature similarity to improve classification in liver and lung cancer. 
MemSAM~\cite{deng2024memsam} utilizes a memory bank of anatomical priors to generate pseudo-masks without manual labels. 
%
Graph-based SSL has also shown promise in classification settings. For instance, GraphX-NET~\cite{aviles2019graphx} models chest X-ray data with graph structures, and NoTeacher~\cite{unnikrishnan2021semi} combines consistency regularization with probabilistic graphical models to impose structural constraints.
While these methods have achieved strong performance in medical diagnosis, \textit{they are predominantly based on black-box models and neglect model transparency and interpretability. }

\vspace{-2mm}
\noindent\textbf{Visual Graph Learning }
combines visual representation with structured relational reasoning and has become an effective paradigm for modeling high-order semantics in vision tasks. 
Unlike traditional CNNs or self-attention models that operate over local pixel or patch-level features, graph-based methods construct relational graphs where each node represents a data instance or image region, and edges encode semantic affinities~\cite{zhu2005semi,chong2020graph,song2022graph}. These approaches allow models to incorporate both explicit and latent relationships, offering a more structured understanding of similarity.
Recent works incorporate hypergraphs to capture richer multi-way relationships beyond pairwise interactions. Hypergraph-based methods~\cite{gao20123,huang2009video} have been applied to GNNs~\cite{han2023vision,srinivas2024vision} for better representation learning. HgVT~\cite{fixelle2025hypergraph} propose dynamic hypergraph construction guided by feature similarity and regularization strategies, enabling structure-aware learning directly within vision transformer architectures. These advances demonstrate the potential of structured learning for enhanced generalization in complex visual recognition tasks. However, none of them explore inter-concept relationships for interpretability.